\documentclass{article}

\usepackage{PRIMEarxiv}

\usepackage[utf8]{inputenc} 
\usepackage[T1]{fontenc}    
\usepackage{hyperref}       
\usepackage{url}            
\usepackage{booktabs}       
\usepackage{amsfonts}       
\usepackage{nicefrac}       
\usepackage{microtype}      
\usepackage{lipsum}
\usepackage{fancyhdr}       
\usepackage{graphicx}       
\graphicspath{{media/}}     
\usepackage{amsmath}
\usepackage{float} 

\title{RLSR: Reinforcement Learning with Supervised Reward Outperforms SFT in Instruction Following
}

\author{
  Zhichao Wang\thanks{\textbf{Corresponding Author}: \texttt{zcwang0201@gmail.com}}, Andy Wong, Ruslan Belkin \\
  Inflection AI \\
}

\begin{document}
\maketitle

\begin{abstract}
After the pretraining stage of LLMs, techniques such as SFT, RLHF, RLVR, and RFT are applied to enhance instruction-following ability, mitigate undesired responses, improve reasoning capability and enable efficient domain adaptation with minimal data. SFT relies on the next-token prediction objective to strengthen instruction following in a base model using a large corpus of human-labeled responses. In contrast, RFT employs a RL-based approach to adapt fine-tuned reasoning models to specific domains with limited supervision. Inspired by RFT, we propose replacing SFT with RLSR to leverage the extensive SFT dataset in an RL framework, thereby improving the base model's instruction-following ability. In RLSR, the base model generates multiple responses for each prompt, and reward scores are computed as the cosine similarity in the semantic embedding space between the generated and human-labeled responses. RLSR can be utilized in multiple ways. It can directly replace SFT, achieving superior performance on instruction-following benchmarks-for example, RLSR (SB) on Qwen-7B (INFINITY) achieved an AlpacaEval win rate of 26.34\%, surpassing SFT's 21.01\%. Furthermore, combining SFT and RLSR further enhances downstream task performance; Qwen-7B (INFINITY) achieved a win rate of 30.73\% when trained with SFT + RLSR.
\end{abstract}


\begin{figure}
    \centering
    \includegraphics[width=0.9\linewidth]{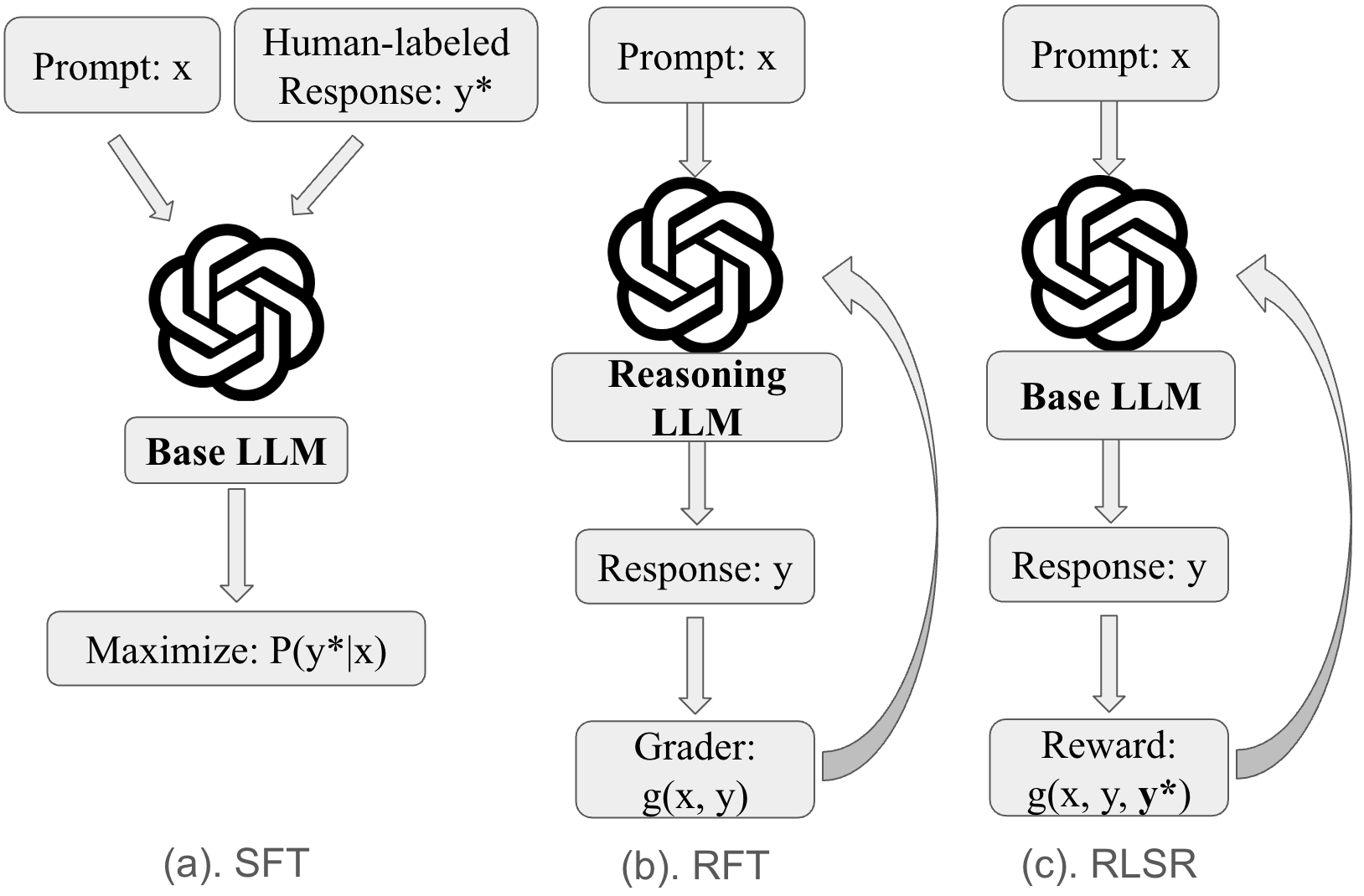}
    \caption{(a) SFT fine-tunes the base model on human-labeled responses to maximize the likelihood of next-token prediction. (b) RFT adapts the reasoning model to specific domains with minimal data with a RL framework, employing a grader function that may optionally incorporate human-labeled responses. (c) RLSR builds upon the RFT framework while leveraging the data used in SFT. It strengthens the base model's instruction-following ability by defining a reward function based on the cosine similarity between the embedding representations of the model-generated and human-labeled responses, utilizing large-scale SFT data.}
    \label{fig:SFT_RFT_RLSR}
\end{figure}

\section{Introduction}
Large Language Models (LLMs) have demonstrated transformative potential across a myriad of tasks following their initial "pretraining stage". However, to become truly capable and safe assistants, they must undergo subsequent post-training to enhance their instruction-following ability, mitigate undesired responses, improve complex reasoning capability, and enable efficient domain adaptation. This typically involves techniques such as Supervised Fine-Tuning (SFT), Reinforcement Learning from Human Feedback (RLHF) \cite{ouyang2022traininglanguagemodelsfollow}, Reinforcement Learning with Verifiable Rewards (RLVR) \cite{shao2024deepseekmathpushinglimitsmathematical}, and Reinforcement Fine-Tuning (RFT) \cite{deepseekai2025deepseekr1incentivizingreasoningcapability}.

The prevailing standard for initial alignment is SFT. SFT leverages a dataset of high-quality, human-labeled $(\text{prompt}, \text{response})$ pairs, optimizing the model with a next-token prediction objective (cross-entropy loss) \cite{brown2020languagemodelsfewshotlearners}. While effective and computationally stable, SFT is a "teacher-forcing" method that strictly enforces human-labeled generations at the token level, inherently limiting the model's "exploratory capacity" and potentially confining it to suboptimal local optima.

In contrast, methods based on Reinforcement Learning (RL), such as RLHF and RLVR, have proven highly effective for complex objectives like safety and reasoning. These approaches shift the objective from token-level match to "maximizing a reward signal", allowing for a crucial balance between exploration and exploitation. RLHF, for instance, trains a Reward Model (RM) on human preference data to guide a subsequent policy optimization step, typically with Proximal Policy Optimization (PPO) \cite{schulman2017proximalpolicyoptimizationalgorithms} or its variants like Group Relative Preference Optimization (GRPO) \cite{shao2024deepseekmathpushinglimitsmathematical}. In RLVR, the reward model is replaced with a verifiable function to improve the model's capability on math and coding tasks. Similarly, RFT adapts a fine-tuned reasoning model to specific domains by utilizing a programmable grader to assign rewards for nuanced objectives.

This disparity raises a critical question: \textbf{Can we replace or augment SFT with a purely RL-based approach that makes full use of the existing, extensive SFT human-labeled dataset to improve the base model's core instruction-following capability?}

Inspired by the success of RL-based methods in enhancing model capabilities beyond SFT's token-level constraints, we propose \textbf{Reinforcement Learning with Supervised Reward (RLSR)}. Borrowing ideas from RFT, RLSR re-frames the SFT process within an RL framework. For each prompt, the base model generates multiple candidate responses. Instead of relying on a learned reward model or sparse correctness signals, RLSR computes a reward score for each candidate based on the cosine similarity in the semantic embedding space between the generated response and the human-labeled response. This reward function directly leverages the high-quality SFT data while introducing the necessary element of exploration inherent to RL. We define the reward $r(x, y, y^*)$ as the cosine similarity between the embeddings $E(x, y)$ of the prompt and generated response $y$ and $E(x, y^*)$ of the prompt and human-labeled response $y^*$. We demonstrate that RLSR can be utilized in two powerful ways:
\begin{enumerate}
    \item \textbf{Direct SFT Replacement}: RLSR can directly substitute the SFT phase, achieving superior performance on instruction-following benchmarks.
    \item \textbf{SFT Enhancement}: RLSR can be applied as a subsequent fine-tuning stage after SFT, creating an \textbf{SFT + RLSR} pipeline that further boosts downstream task performance.
\end{enumerate}

Our extensive experiments across three model sizes (Llama-8B, Qwen-7B, Qwen-32B) and two high-quality SFT datasets (TULU and INFINITY) reveal the significant advantage of our RLSR method. We show that RLSR consistently outperforms the traditional SFT method on key instruction-following and generation metrics:
\begin{itemize}
    \item RLSR Outperforms SFT: On the Qwen-7B model fine-tuned with the INFINITY dataset, RLSR (SB) achieved an AlpacaEval win rate of 26.34\%, significantly surpassing the SFT win rate of 21.01\%. This indicates a superior ability to follow open-ended instructions as judged by a leading evaluation benchmark.
    \item SFT+RLSR is the Strongest Approach: Combining the methods further maximizes performance. The SFT + RLSR pipeline pushed the Qwen-7B (INFINITY) model to its highest performance, achieving an AlpacaEval win rate of 30.73\%.
\end{itemize}
These results confirm our hypothesis: by leveraging the supervisory signal of SFT data within an RL framework, \textbf{RLSR significantly enhances the LLM's instruction-following capability} beyond the limits of token-level supervision.

\section{Methodology}
\subsection{SFT}

While the pretrained base models exhibit strong generative capabilities, they often fail to reliably follow instructions-frequently producing repetitive or input-extending outputs. To better align these models with instruction-following and question-answering tasks, a SFT stage is introduced.  

In SFT, the base model is trained on a dataset of prompt–response pairs, where each prompt is denoted by $x$ and its corresponding target response by $y$. The model defines a conditional probability distribution $\pi_\theta(y|x)$, representing the likelihood of generating response $y$ given input $x$. The objective of SFT is to optimize the model parameters $\theta$ by maximizing this conditional likelihood-equivalently, minimizing the negative log-likelihood or cross-entropy loss as formulated in Eq.~\ref{eq: SFT loss}, where $\{y_1, \cdots, y_N\}$ denote the $N$ tokens in the response sequence:

\begin{equation}
\label{eq: SFT loss}
L_{\text{SFT}}(\pi_\theta) = -\log \pi_\theta(y|x) = - \sum_{i=1}^{N} \log \pi_\theta(y_i \mid x, y_{<i})
\end{equation}

Empirical studies have demonstrated that SFT substantially enhances an LLM's ability to follow user instructions and generate contextually appropriate answers. However, because SFT relies on a \textit{teacher-forcing} paradigm-where the model is always conditioned on the ground-truth sequence during training-it limits the model's capacity for autonomous exploration and generalization. Moreover, even after SFT, undesired behaviors such as biased or unsafe generations may still persist.

\subsection{RLHF, RLVR and RFT}

To mitigate the generation of biased or toxic content in the instruct model from SFT, RLHF has been proposed. RLHF consists of two main stages: (1) RM training and (2) reinforcement learning fine-tuning. In the first stage, the reward model is trained on pairwise human preference data, where each sample consists of a prompt $x$ and two responses $y_w$ (desired) and $y_l$ (undesired), following the Bradley–Terry (BT) model. The content of RM deviates from the main topic of this paper and more details can be found in \cite{ouyang2022traininglanguagemodelsfollow}.

Once the reward model is trained, it is employed to guide an online reinforcement learning process. The policy $\pi_\theta$ is optimized to maximize the expected reward while remaining close to a reference policy $\pi_{\text{ref}}$, which reflects the pretrained distribution of the base model. This trade-off between reward maximization and policy stability is expressed as:

\begin{equation}
\label{eq:RL_objective}
\pi^*_\theta(y|x) = \arg\max_{\pi_\theta}
\mathbb{E}_{x \sim D} \Big[
\mathbb{E}_{y \sim \pi_\theta(y|x)} [r_\phi(x, y)]
- \beta D_{\text{KL}}\big(\pi_\theta(y|x)\,\|\,\pi_{\text{ref}}(y|x)\big)
\Big]
\end{equation}

where $\beta$ controls the strength of the KL-divergence penalty. Proximal Policy Optimization (PPO) is typically used to solve this objective. The objective in Equation \ref{eq:RL_objective} is shared among RLHF, RLVR, RFT and RLSR and the only difference is the definition of the reward model.

Although RLHF is effective, it demands substantial computational resources-it requires maintaining separate reward, value, and policy networks-and often suffers from instability during training. To alleviate these challenges, GRPO has been proposed as a simplified alternative to PPO. Unlike PPO's actor–critic framework, which trains a value function using cumulative rewards from the RM, GRPO replaces the learned value function with the mean reward computed across multiple sampled responses. This modification eliminates the need for a separate value model, thereby reducing FLOPs, memory usage, and training complexity.

LLMs fine-tuned with RLHF and GRPO exhibit a strong ability to suppress undesired or harmful generations. However, these methods still struggle on complex reasoning tasks such as mathematics or programming. To address this limitation, RLVR has been introduced. Unlike RLHF, which relies on a learned reward model, RLVR uses verifiable, automatically computed signals-such as correctness of a mathematical answer or whether generated code passes test cases-as rewards. Although these rewards are sparse, they are precise and directly aligned with task success. The combination of RLVR and GRPO, as implemented in \textit{DeepSeek R1}~\cite{deepseekai2025deepseekr1incentivizingreasoningcapability}, substantially improves an LLM's reasoning ability, enabling longer and more coherent chains of thought (CoT) and fostering “aha moments,” where the model recognizes and corrects prior reasoning errors.

While RLHF primarily focuses on mitigating harmful outputs and RLVR enhances reasoning performance, RFT aims to adapt an OpenAI reasoning model (e.g., \texttt{o4-mini}) to expert-level performance in domain-specific applications \cite{OpenAI_RFT}. Instead of relying on fixed “correct” answers as in SFT, RFT employs a programmable grader function that evaluates model outputs using user-defined feedback signals. Typical components of the grader include (1) string matching, (2) semantic similarity, (3) score-model grading, (4) label-model grading, and (5) Python-based code execution, which optionally utilizes the human-labeled responses. The model is trained to prioritize high-scoring outputs, aligning generation behavior with nuanced objectives such as style, safety, or domain accuracy. This paradigm is particularly valuable in low-resource or domain-shift settings where human-labeled data are scarce and traditional fine-tuning methods generalize poorly.

\subsection{RLSR}

From previous studies, it can be observed that RL-based methods, which incorporate a balance between exploration and exploitation, substantially enhance the capabilities of LLMs. However, the SFT stage remains a purely teacher-forcing process, involving only exploitation without exploration. This naturally raises a question: \textbf{Can the SFT data itself be leveraged within an RL framework to further improve model performance?}

Motivated by the question of how to effectively leverage SFT data within an RL framework, and inspired by the RFT paradigm, we propose RLSR, a method that fine-tunes the base model using reinforcement learning while fully utilizing the existing SFT dataset. The key component of RLSR lies in the design of its reward function, which is derived directly from the SFT prompt–response pairs labeled by humans. In line with the principle of SFT, the reward reflects the semantic similarity between the model-generated response and the human-labeled response, under the assumption that semantically closer responses are of higher quality and more desirable. Following the embedding-based approaches adopted in DSSM \cite{huang2013learning}, CLIP \cite{radford2021learningtransferablevisualmodels}, and RAG \cite{lewis2021retrievalaugmentedgenerationknowledgeintensivenlp}, we define the reward as the cosine similarity between the embeddings of the generated response $y^i$ and the human-labeled response $y^*$. Let $E(\cdot)$ denote the text encoder that extracts embeddings from responses. The reward function is formulated as:

\begin{equation}
r(x, y^i, y^*) = \frac{E(y^i) \cdot E(y^*)}{||E(y^i)||_2 \cdot ||E(y^*)||_2}
\label{RLSR:reward-cosine}
\end{equation}

Here, $E(y^i)$ and $E(y^*)$ represent the embeddings of the generated and human-labeled responses, respectively. The cosine similarity ensures that semantically aligned responses yield higher rewards, which are subsequently used in the reinforcement learning process to fine-tune the language model.

There are two primary strategies to integrate RLSR. The first approach replaces SFT entirely with RLSR, and empirical results indicate that RLSR consistently outperforms SFT across various models, datasets, and benchmarks. The second approach introduces RLSR as an intermediate stage between SFT and RLHF/RLVR, resulting in the following training pipeline:
\begin{enumerate}
    \item Supervised Fine-Tuning (SFT),
    \item Reinforcement Learning with Supervised Reward (RLSR),
    \item Reinforcement Learning from Human Feedback or Verifiable Reward (RLHF/RLVR).
\end{enumerate}

Finally, we summarize the advantages and trade-offs between RLSR and SFT. The primary strengths of SFT lie in its efficiency and stability-by using a teacher-forcing paradigm, the model directly learns the correct next token, which accelerates convergence but eliminates exploration. In contrast, RLSR introduces exploration by evaluating diverse candidate responses through reward-guided optimization. Although this process consumes more computational resources (FLOPs), it often leads to better overall performance when appropriate reward signals are designed.

\subsection{Differences between RLSR and RFT}

It is essential to distinguish between RLSR and RFT across four key dimensions: \textbf{model}, \textbf{goal}, \textbf{data scale}, and \textbf{reward}. 
From the \textbf{model} perspective, RFT operates on a reasoning model, whereas RLSR fine-tunes the base model directly. 
From the \textbf{goal} perspective, RFT aims to efficiently adapt the model to new domains with minimal data, while RLSR focuses on enhancing the instruction-following capability of the base model. 
Regarding the \textbf{data scale}, RFT seeks to minimize data usage due to the scarcity of domain-specific data, whereas RLSR leverages the full SFT dataset, which is typically large in scale. 
Finally, from the \textbf{reward} perspective, RFT employs a grader function-either rule-based or model-based-that may optionally incorporate human-labeled responses. 
In contrast, RLSR defines its reward based on the cosine similarity between the embeddings of generated responses and human-labeled responses, aligning the base model's behavior with human preferences through explicit supervision.

\section{Experiments}
This section presents the experimental setup used to evaluate the effectiveness of RLSR compared to SFT in fine-tuning base models for instruction-following tasks. We describe the models, datasets, and evaluation protocols employed in our study.

\subsection{Models, Datasets, and Evaluation}
We conduct experiments using three representative base models: (1) \texttt{meta-llama/Llama-3.1-8B} \cite{grattafiori2024llama3herdmodels}, (2) \texttt{Qwen/Qwen2.5-7B} \cite{qwen2025qwen25technicalreport}, and (3) \texttt{Qwen/Qwen2.5-32B} \cite{qwen2025qwen25technicalreport}. These models are chosen to represent two major open-source LLM families-\textbf{LLaMA} and \textbf{Qwen}-providing diversity in both architectural design and model scale. For consistency in prompt formatting, their instruction-tuned variants-\texttt{Llama-3.1-8B-Instruct}, \texttt{Qwen2.5-7B-Instruct}, and \texttt{Qwen2.5-32B-Instruct}-are adopted.

Two embedding models are employed as reward functions: \texttt{sentence-transformers/all-MiniLM-L6-v2} (SentenceBERT, abbreviated as SB) \cite{edoardo_federici_2022} and \texttt{Qwen/Qwen3-Embedding-8B} (Qwen-EM) \cite{qwen3embedding}. The SB model is relatively lightweight, containing 33 million parameters and supporting a 512-token context window. During embedding generation, inputs are truncated to the first 512 tokens-sufficient for most outputs. In contrast, the Qwen model is substantially larger, with 8 billion parameters and a 32k-token context window, enabling it to encode entire sequences without truncation. To clearly distinguish between the base and embedding models, we denote the base models as Qwen-7B and Qwen-32B, and refer to the embedding model simply as Qwen-EM.

\begin{table}[!htb]
\caption{Fine-tune Llama-8B, Qwen-7B and Qwen-32B utilizing SFT, RLSR and SFT+RLSR with TULU and INFINITY datasets and their performances on truthfulqa, hendrycks\_math, humaneval, bbq, winogrande and mmlu\_pro.}
\label{tab:performance_part_1}
\begin{tabular}{lcccccc}
\hline
\hline
\textbf{Model (TULU)} & \textbf{truthfulqa} & \textbf{hendrycks\_math} & \textbf{humaneval} & \textbf{bbq} & \textbf{winogrande} & \textbf{mmlu\_pro} \\
\hline
\hline
Llama & 45.23 & \textbf{14.00} & 37.20 & 43.89 & 77.19 & 32.95 \\
SFT & 49.22 & 10.12 & \textbf{44.51} & \textbf{49.19} & 76.64 & 23.11 \\
RLSR (SB) & \textbf{52.50} & 13.82 & 40.24 & 46.75 & 76.64 & 32.72 \\
RLSR (Qwen-EM) & 50.83 & 13.62 & 42.07 & 47.05 & \textbf{78.14} & \textbf{34.15} \\
\hline
SFT+RLSR (SB) & 50.42 & 10.84 & 46.34 & 50.38 & \textbf{71.98} & 24.26 \\
SFT+RLSR (Qwen-EM) & \textbf{51.11} & \textbf{11.28} & \textbf{47.56} & \textbf{50.75} & 71.82 & \textbf{24.79} \\
\hline
\hline
Qwen7B & 56.37 & \textbf{25.00} & 56.71 & 49.31 & \textbf{75.69} & 43.97 \\
SFT & 48.86 & 20.00 & \textbf{65.24} & 47.22 & 74.43 & 41.62 \\
RLSR (SB) & 53.83 & 24.30 & 56.71 & \textbf{50.39} & 74.35 & \textbf{44.31} \\
RLSR (Qwen-EM) & \textbf{57.91} & 24.88 & 62.80 & 49.54 & 74.35 & 44.10 \\
\hline
SFT+RLSR (SB) & 50.40 & 17.36 & \textbf{66.46} & 49.52 & \textbf{71.90} & 40.88 \\
SFT+RLSR (Qwen-EM) & \textbf{52.15} & \textbf{18.30} & 64.63 & \textbf{52.20} & 69.14 & \textbf{41.62} \\
\hline
\hline
Qwen32B & 57.76 & \textbf{35.78} & 48.17 & 51.44 & 81.85 & 58.14 \\
SFT & 48.60 & 33.24 & \textbf{65.24} & 51.72 & 80.66 & 55.39 \\
RLSR (SB) & 57.79 & 33.02 & 54.27 & \textbf{51.86} & 80.82 & 58.08 \\
RLSR (Qwen-EM) & \textbf{58.67} & 34.24 & 59.15 & 51.64 & \textbf{82.32} & \textbf{58.21} \\
\hline
SFT+RLSR (SB) & 49.16 & 30.68 & 62.80 & 54.72 & 78.22 & 55.81 \\
SFT+RLSR (Qwen-EM) & \textbf{49.64} & \textbf{31.66} & \textbf{68.90} & \textbf{55.15} & \textbf{78.53} & \textbf{56.86} \\
\hline
\hline
\textbf{Model (INFINITY)} & \textbf{truthfulqa} & \textbf{hendrycks\_math} & \textbf{humaneval} & \textbf{bbq} & \textbf{winogrande} & \textbf{mmlu\_pro} \\
\hline
\hline
Llama & 45.23 & 14.00 & 37.20 & 43.89 & \textbf{77.19} & 32.95 \\
SFT & 51.85 & 12.20 & 34.15 & \textbf{48.22} & 74.27 & 29.49 \\
RLSR (SB) & 52.29 & 14.02 & \textbf{39.63} & 44.95 & 76.80 & \textbf{33.72} \\
RLSR (Qwen-EM) & \textbf{52.71} & \textbf{14.30} & 39.02 & 45.64 & 76.48 & 33.10 \\
\hline
SFT+RLSR (SB) & 52.60 & \textbf{13.02} & 28.05 & 49.57 & \textbf{72.69} & 30.08 \\
SFT+RLSR (Qwen-EM) & \textbf{53.64} & 12.92 & \textbf{43.29} & \textbf{50.22} & 71.35 & \textbf{30.11} \\
\hline
\hline
Qwen7B & 56.37 & \textbf{25.00} & 56.71 & 49.31 & 75.69 & 43.97 \\
SFT & 54.26 & 24.56 & \textbf{63.41} & 46.12 & 75.06 & 43.85 \\
RLSR (SB) & \textbf{59.06} & \textbf{25.00} & 57.93 & 49.35 & \textbf{75.77} & \textbf{44.07} \\
RLSR (Qwen-EM) & 58.20 & 24.96 & 60.37 & \textbf{49.84} & 74.66 & 43.89 \\
\hline
SFT+RLSR (SB) & 56.60 & 23.42 & \textbf{68.90} & 45.89 & 73.72 & 43.79 \\
SFT+RLSR (Qwen-EM) & \textbf{57.18} & \textbf{23.64} & 65.24 & \textbf{47.46} & \textbf{74.66} & \textbf{44.14} \\
\hline
\hline
Qwen32B & 57.76 & 35.78 & 48.17 & 51.44 & 81.85 & 58.14 \\
SFT & 58.45 & \textbf{36.60} & \textbf{57.93} & \textbf{51.74} & \textbf{82.00} & 56.22 \\
RLSR (SB) & \textbf{59.40} & 32.18 & 52.44 & 51.64 & 81.53 & \textbf{58.49} \\
RLSR (Qwen-EM) & 59.33 & 34.86 & 53.05 & 51.73 & 81.93 & 58.14 \\
\hline
SFT+RLSR (SB) & 55.27 & \textbf{37.12} & 57.93 & 51.80 & 80.66 & 56.42 \\
SFT+RLSR (Qwen-EM) & \textbf{55.62} & 36.64 & \textbf{60.98} & \textbf{52.16} & \textbf{81.22} & \textbf{56.65} \\
\hline
\hline
\end{tabular}
\end{table}

\begin{table}[!htb]
\caption{Fine-tune Llama-8B, Qwen-7B and Qwen-32B utilizing SFT, RLSR and SFT+RLSR with TULU and INFINITY datasets and their performances on gsm8k, gpqa, mmlu, mbpp, arc\_challenge and toxigen.}
\label{tab:performance_part_2}
\begin{tabular*}{\textwidth}{l @{\extracolsep{\fill}} cccccc}
\hline
\hline
\textbf{Model (TULU)} & \textbf{gsm8k} & \textbf{gpqa} & \textbf{mmlu} & \textbf{mbpp} & \textbf{arc\_challenge} & \textbf{toxigen} \\
\hline
\hline
Llama & 49.81 & \textbf{31.88} & 65.40 & 48.6 & 54.52 & 42.66 \\
SFT & 58.61 & 27.94 & 57.01 & 42.2 & 54.27 & \textbf{56.70} \\
RLSR (SB) & \textbf{59.67} & 31.04 & 65.03 & \textbf{50.2} & 57.51 & 43.30 \\
RLSR (Qwen-EM) & 58.53 & 30.70 & \textbf{65.45} & 49.6 & \textbf{59.13} & 44.26 \\
\hline
SFT+RLSR (SB) & \textbf{68.16} & 28.02 & 57.61 & \textbf{45.6} & \textbf{55.63} & \textbf{50.85} \\
SFT+RLSR (Qwen-EM) & \textbf{68.16} & \textbf{28.69} & \textbf{57.74} & 44.0 & 55.29 & 49.26 \\
\hline
\hline
Qwen7B & 81.50 & \textbf{32.63} & 74.24 & 63.8 & 59.39 & 57.13 \\
SFT & 73.46 & 31.96 & 72.26 & 61.0 & 58.11 & 56.70 \\
RLSR (SB) & 83.17 & 31.12 & 74.23 & \textbf{66.6} & 61.35 & 56.91 \\
RLSR (Qwen-EM) & \textbf{84.38} & 31.54 & \textbf{74.37} & 65.6 & \textbf{62.37} & \textbf{57.45} \\
\hline
SFT+RLSR (SB) & 76.27 & 32.47 & \textbf{72.16} & 60.2 & 59.04 & \textbf{57.13} \\
SFT+RLSR (Qwen-EM) & \textbf{77.03} & \textbf{32.97} & 71.99 & \textbf{62.8} & \textbf{59.22} & 56.81 \\
\hline
\hline
Qwen32B & 84.61 & 39.35 & 83.29 & 73.2 & 67.06 & 81.49 \\
SFT & 87.95 & 39.51 & 83.03 & 73.6 & 65.10 & 54.89 \\
RLSR (SB) & \textbf{91.66} & \textbf{39.68} & 83.11 & 74.4 & 69.03 & 81.06 \\
RLSR (Qwen-EM) & 90.75 & 39.60 & \textbf{83.41} & \textbf{76.4} & \textbf{69.80} & \textbf{86.17} \\
\hline
SFT+RLSR (SB) & \textbf{90.52} & 40.10 & 82.77 & \textbf{74.2} & \textbf{68.26} & \textbf{71.91} \\
SFT+RLSR (Qwen-EM) & 89.99 & \textbf{40.52} & \textbf{82.89} & 73.4 & 67.92 & 70.74 \\
\hline
\hline
\textbf{Model (INFINITY)} & \textbf{gsm8k} & \textbf{gpqa} & \textbf{mmlu} & \textbf{mbpp} & \textbf{arc\_challenge} & \textbf{toxigen} \\
\hline
\hline
Llama & 49.81 & 31.88 & \textbf{65.40} & 48.6 & 54.52 & 42.66 \\
SFT & 51.10 & 28.44 & 61.47 & 43.0 & 54.86 & 42.55 \\
RLSR (SB) & 57.16 & 29.61 & 65.05 & \textbf{52.2} & \textbf{56.31} & \textbf{45.11} \\
RLSR (Qwen-EM) & \textbf{58.83} & \textbf{32.63} & 65.35 & 51.4 & 56.14 & 43.30 \\
\hline
SFT+RLSR (SB) & 54.51 & 27.60 & \textbf{61.73} & 41.8 & \textbf{56.48} & 42.45 \\
SFT+RLSR (Qwen-EM) & \textbf{55.04} & \textbf{27.77} & 61.72 & \textbf{43.4} & 56.14 & \textbf{42.55} \\
\hline
\hline
Qwen7B & 81.50 & 32.63 & 74.24 & 63.8 & 59.39 & \textbf{57.13} \\
SFT & 78.54 & 32.21 & 72.86 & 65.0 & 56.74 & 55.96 \\
RLSR (SB) & \textbf{83.47} & 32.05 & \textbf{74.41} & \textbf{66.6} & 59.81 & 56.91 \\
RLSR (Qwen-EM) & \textbf{83.47} & \textbf{32.89} & 74.20 & 66.2 & \textbf{62.37} & 56.91 \\
\hline
SFT+RLSR (SB) & \textbf{84.23} & \textbf{33.14} & 72.74 & \textbf{63.6} & 59.90 & \textbf{56.81} \\
SFT+RLSR (Qwen-EM) & \textbf{84.23} & 32.97 & \textbf{72.84} & 63.2 & \textbf{61.09} & 56.17 \\
\hline
\hline
Qwen32B & \textbf{84.61} & \textbf{39.35} & \textbf{83.29} & 73.2 & 67.06 & 81.49 \\
SFT & 82.87 & 37.67 & 83.21 & 73.6 & 68.09 & 67.02 \\
RLSR (SB) & 81.96 & 39.01 & 83.21 & \textbf{75.6} & 67.49 & 77.66 \\
RLSR (Qwen-EM) & 83.40 & 38.84 & 83.20 & 75.0 & \textbf{68.94} & \textbf{86.38} \\
\hline
SFT+RLSR (SB) & \textbf{84.15} & 37.08 & \textbf{83.20} & 74.2 & 69.28 & 79.68 \\
SFT+RLSR (Qwen-EM) & 82.87 & \textbf{37.92} & 82.87 & \textbf{77.0} & \textbf{70.48} & \textbf{83.19} \\
\hline
\hline
\end{tabular*}
\end{table}

\begin{table}[!htb]
\caption{Fine-tune Llama-8B, Qwen-7B and Qwen-32B utilizing SFT, RLSR and SFT+RLSR with TULU and INFINITY datasets and their performances on math\_hard, ifeval, hellaswag, musr, winogender and bbh.}
\label{tab:performance_part_3}
\begin{tabular*}{\textwidth}{l @{\extracolsep{\fill}} cccccc}
\hline
\hline
\textbf{Model (TULU)} & \textbf{math\_hard} & \textbf{ifeval} & \textbf{hellaswag} & \textbf{musr} & \textbf{winogender} & \textbf{bbh} \\
\hline
\hline
Llama & 5.51 & 17.99 & 61.51 & 38.10 & 61.11 & 46.22 \\
SFT & \textbf{7.55} & \textbf{58.39} & 59.11 & 40.61 & \textbf{62.50} & 44.28 \\
RLSR (SB) & 7.10 & 20.62 & \textbf{63.88} & 42.33 & 61.53 & 47.68 \\
RLSR (Qwen-EM) & 5.44 & 13.67 & 63.64 & \textbf{47.09} & 61.11 & \textbf{48.05} \\
\hline
SFT+RLSR (SB) & \textbf{9.59} & 62.71 & \textbf{60.88} & \textbf{43.92} & 63.75 & 45.79 \\
SFT+RLSR (Qwen-EM) & 8.99 & \textbf{65.83} & 60.28 & 42.86 & \textbf{64.03} & \textbf{46.09} \\
\hline
\hline
Qwen7B & 22.81 & 41.97 & 60.32 & 43.78 & 58.61 & \textbf{53.84} \\
SFT & 24.70 & \textbf{56.71} & 58.95 & 44.71 & 56.81 & 51.15 \\
RLSR (SB) & 26.06 & 53.60 & \textbf{61.40} & \textbf{45.50} & 60.00 & 52.65 \\
RLSR (Qwen-EM) & \textbf{27.72} & 55.64 & 61.22 & 43.52 & \textbf{60.42} & 52.06 \\
\hline
SFT+RLSR (SB) & 27.19 & 66.67 & 61.46 & \textbf{46.16} & 58.19 & 51.71 \\
SFT+RLSR (Qwen-EM) & \textbf{29.61} & \textbf{67.51} & \textbf{61.80} & 45.63 & \textbf{58.33} & \textbf{52.92} \\
\hline
\hline
Qwen32B & 36.25 & 49.16 & 65.87 & 48.81 & 60.14 & \textbf{67.44} \\
SFT & 39.73 & \textbf{68.23} & 65.05 & 46.83 & 61.53 & 65.53 \\
RLSR (SB) & 39.27 & 60.31 & \textbf{67.23} & 49.87 & 61.25 & 66.50 \\
RLSR (Qwen-EM) & \textbf{41.09} & 65.47 & 67.05 & \textbf{50.26} & \textbf{61.94} & 66.36 \\
\hline
SFT+RLSR (SB) & \textbf{44.94} & 75.54 & 68.16 & \textbf{49.47} & \textbf{61.39} & 66.13 \\
SFT+RLSR (Qwen-EM) & 43.20 & \textbf{81.53} & \textbf{69.03} & 48.41 & 61.25 & \textbf{66.46} \\
\hline
\hline
\textbf{Model (INFINITY)} & \textbf{math\_hard} & \textbf{ifeval} & \textbf{hellaswag} & \textbf{musr} & \textbf{winogender} & \textbf{bbh} \\
\hline
\hline
Llama & 5.51 & 17.99 & 61.51 & 38.10 & \textbf{61.11} & 46.22 \\
SFT & 5.29 & \textbf{46.16} & 59.55 & 41.27 & 60.69 & 46.14 \\
RLSR (SB) & \textbf{7.02} & 17.39 & 62.58 & \textbf{46.03} & \textbf{61.11} & 47.54 \\
RLSR (Qwen-EM) & 6.34 & 17.87 & \textbf{62.96} & 42.86 & \textbf{61.11} & \textbf{47.79} \\
\hline
SFT+RLSR (SB) & \textbf{7.63} & 46.52 & 60.37 & \textbf{46.56} & 61.25 & 47.08 \\
SFT+RLSR (Qwen-EM) & 6.34 & \textbf{47.24} & \textbf{60.64} & 44.84 & \textbf{61.94} & \textbf{47.91} \\
\hline
\hline
Qwen7B & 22.81 & 41.97 & 60.32 & 43.78 & 58.61 & 53.84 \\
SFT & 21.98 & 50.36 & 59.36 & \textbf{45.77} & 57.22 & \textbf{54.30} \\
RLSR (SB) & 26.51 & 55.52 & 61.21 & 41.93 & \textbf{60.00} & 53.71 \\
RLSR (Qwen-EM) & \textbf{26.96} & \textbf{57.43} & \textbf{61.62} & 43.52 & 59.17 & 53.55 \\
\hline
SFT+RLSR (SB) & 21.30 & 54.68 & 60.39 & \textbf{46.43} & 59.17 & \textbf{53.79} \\
SFT+RLSR (Qwen-EM) & \textbf{21.68} & \textbf{56.35} & \textbf{60.86} & 44.71 & \textbf{59.44} & \textbf{53.79} \\
\hline
\hline
Qwen32B & 36.25 & 49.16 & 65.87 & 48.81 & 60.14 & 67.44 \\
SFT & 36.56 & 64.27 & 65.13 & 47.49 & 59.86 & \textbf{69.50} \\
RLSR (SB) & 38.97 & \textbf{67.15} & 66.91 & 47.09 & 59.58 & 67.11 \\
RLSR (Qwen-EM) & \textbf{39.20} & 66.79 & \textbf{67.15} & \textbf{49.07} & \textbf{64.58} & 66.83 \\
\hline
SFT+RLSR (SB) & \textbf{36.78} & 68.23 & 66.33 & \textbf{49.07} & 58.47 & \textbf{69.14} \\
SFT+RLSR (Qwen-EM) & 36.03 & \textbf{68.47} & \textbf{66.46} & 47.88 & \textbf{61.25} & 67.82 \\
\hline
\hline
\end{tabular*}
\end{table}

Two instruction-tuning datasets, \textbf{TULU} \cite{lambert2025tulu3pushingfrontiers} and \textbf{INFINITY} \cite{li2025infinityinstructscalinginstruction}, each containing 100k sampled single-turn examples, are employed as the high-quality instruction-following data. \textbf{RLSR} is implemented using \textbf{VERL} \cite{sheng2024hybridflow}, while \textbf{SFT} is conducted with \textbf{OpenRLHF} \cite{hu2025openrlhfeasytousescalablehighperformance} because VERL's SFT module is unstable and generating repetitive responses. Learning rates are empirically tuned for stability---set to $1\times10^{-6}$ for \textbf{LLaMA} and $3\times10^{-6}$ for both \textbf{Qwen} models. During rollout, the maximum sequence length is capped at 4096 tokens and the number of rollout for each prompt is 8, and the KL coefficient in the reward function is set to 0.001. The training batch size for training is 1024 and the PPO mini batch size is 256. For SFT training, a uniform learning rate of $3\times10^{-5}$ is applied across all models and the training batch size is 256.

In addition, we introduce a specific technique to mitigate repetitive generation. The Longest Common Substring (LCS) algorithm is employed to detect the longest repeated segment within the generated response. A penalty reward of $-1$ is applied if the following two conditions are satisfied: (1) $LCS(y) > 100$, meaning the detected substring exceeds 100 characters in length; and (2) $\frac{LCS(y)}{|y^*|} > 0.1$, indicating that the repeated segment accounts for more than 10\% of the human-labeled reference response. Although we observe that the training process remains stable even without this penalty, we include it as a precautionary measure to further reduce redundancy.

Model performance is assessed using \textbf{18 standardized benchmarks} from the \textbf{LM Evaluation Harness} \cite{eval-harness}, encompassing reasoning, mathematics, programming, bias, and factuality. These benchmarks include \textbf{truthfulQA} \cite{lin2022truthfulqameasuringmodelsmimic}, \textbf{hendrycks\_Math} \cite{hendrycksmath2021}, \textbf{humanEval} \cite{chen2021evaluating}, \textbf{BBQ} \cite{parrish-etal-2022-bbq}, \textbf{Winogrande} \cite{ai2:winogrande}, \textbf{MMLU-Pro} \cite{wang2024mmlu}, \textbf{GSM8K} \cite{cobbe2021gsm8k}, \textbf{GPQA} \cite{rein2023gpqagraduatelevelgoogleproofqa}, \textbf{MMLU} \cite{hendrycks2021ethics}, \textbf{MBPP} \cite{austin2021program}, \textbf{ARC-Challenge} \cite{clark2018thinksolvedquestionanswering}, \textbf{ToxiGen} \cite{hartvigsen2022toxigenlargescalemachinegenerateddataset}, \textbf{Math\_Hard} \cite{fan2024hardmathbenchmarkdatasetchallenging}, \textbf{IFEval} \cite{zhou2023instructionfollowingevaluationlargelanguage}, \textbf{HellaSwag} \cite{zellers2019hellaswag}, \textbf{MuSR} \cite{sprague2024musrtestinglimitschainofthought}, \textbf{Winogender} \cite{ai2:winogrande}, and \textbf{BBH} \cite{suzgun2022challengingbigbenchtaskschainofthought}. In addition, \textbf{AlpacaEval} \cite{dubois2024length} and \textbf{Arena-Hard} \cite{arenahard2024} are employed to evaluate instruction-following ability and open-ended generation quality using human-preference metrics.

\begin{table}[!htb]
\caption{Fine-tune Llama-8B, Qwen-7B and Qwen-32B utilizing SFT, RLSR and SFT+RLSR with TULU and INFINITY datasets and their performances on generation tasks including Alpaca-eval and Arena-hard.}
\centering
\begin{tabular*}{\textwidth}{l @{\extracolsep{\fill}} ccccc}
\hline
\hline
\textbf{Model Name (TULU)} & \multicolumn{3}{c}{\textbf{Alpaca-eval}} & \multicolumn{2}{c}{\textbf{Arena-hard}} \\ 
\cmidrule(lr){2-4} \cmidrule(lr){5-6}
 & \textbf{LC WR} & \textbf{WR} & \textbf{Length} & \textbf{Creative Writing} & \textbf{Style Control}  \\ 
\hline
\hline
Llama  & 0.16 & 0.50 & 26514 & 0.1 & 0.7 \\
SFT  & 4.08 & 3.82 & 2205 & 0.3 & 1.1 \\
RLSR (SB) & \textbf{8.61} & \textbf{6.22} & 1177 & 0.9 & \textbf{1.5} \\
RLSR (Qwen-EM)  & 7.22 & 4.89 & 1118 & \textbf{1.2} & 1.3 \\
\hline
SFT+RLSR (SB)  & \textbf{10.82} & \textbf{9.09} & 1460 & \textbf{1.3} & \textbf{1.6} \\
SFT+RLSR (Qwen-EM)  & 10.56 & 7.27 & 1077 & 1.2 & 1.4 \\
\hline
\hline
Qwen-7B  & 8.98 & \textbf{9.70} & 7021 & 0.9 & 1.4 \\
SFT  & 8.28 & 6.74 & 1508 & 1.0 & 1.6 \\
RLSR (SB) & \textbf{16.43} & 8.14 & 933 & \textbf{1.9} & 2.1 \\
RLSR (Qwen-EM)  & 14.23 & 7.21 & 940 & 1.2 & \textbf{2.3} \\
\hline
SFT+RLSR (SB)  & 15.72 & 8.84 & 1152 & \textbf{1.5} & \textbf{2.2} \\
SFT+RLSR (Qwen-EM)  & \textbf{17.49} & \textbf{9.66} & 1064 & 1.4 & 2.0 \\
\hline
\hline
Qwen-32B  & 5.97 & 7.11 & 14293 & 2.0 & 2.1 \\
SFT  & 14.09 & \textbf{10.85} & 1350 & 0.6 & 2.9 \\
RLSR (SB) & 19.11 & 10.33 & 983 & 2.5 & 3.2 \\
RLSR (Qwen-EM)  & \textbf{20.13} & 10.36 & 968 & \textbf{2.9} & \textbf{4.2} \\
\hline
SFT+RLSR (SB)  & 19.43 & 13.10 & 1152 & 2.8 & \textbf{4.0} \\
SFT+RLSR (Qwen-EM)  & \textbf{20.51} & \textbf{13.11} & 1055 & \textbf{3.5} & 3.9 \\
\hline
\hline
\textbf{Model Name (INFINITY)} & \multicolumn{3}{c}{\textbf{Alpaca-eval}} & \multicolumn{2}{c}{\textbf{Arena-hard}} \\ 
\cmidrule(lr){2-4} \cmidrule(lr){5-6}
 & \textbf{LC WR} & \textbf{WR} & \textbf{Length} & \textbf{Creative Writing} & \textbf{Style Control}  \\ 
\hline
\hline
Llama  & 0.16 & 0.50 & 26514 & 0.1 & 0.7 \\
SFT  & \textbf{14.92} & \textbf{15.23} & 3370 & 0.7 & 1.2 \\
RLSR (SB) & 11.17 & 8.97 & 1504 & \textbf{1.2} & \textbf{1.5} \\
RLSR (Qwen-EM)  & 10.83 & 10.04 &1759  & 0.9 & 1.3 \\
\hline
SFT+RLSR (SB)  & 24.70 & 21.36 & 1845 & 2.1 & \textbf{1.4} \\
SFT+RLSR (Qwen-EM)  & \textbf{30.51} & \textbf{29.51} & 1977 & \textbf{2.8} & \textbf{1.4} \\
\hline
\hline
Qwen-7B  & 8.98 & 9.70 & 7021 & 0.9 & 1.4 \\
SFT  & 21.01 & 19.56 & 2164 & 0.8 & 1.6 \\
RLSR (SB) & 26.34 & 22.57 & 1785 & 2.8 & \textbf{2.3} \\
RLSR (Qwen-EM)  & \textbf{30.36} & \textbf{34.96} & 2269 & \textbf{3.0} & 2.2 \\
\hline
SFT+RLSR (SB)  & 30.73 & 22.64 & 1576 & 3.1 & 2.1 \\
SFT+RLSR (Qwen-EM)  & \textbf{31.85} & \textbf{27.56} & 1797 & \textbf{3.6} &  \textbf{2.5} \\
\hline
\hline
Qwen-32B  & 5.97 & 7.11 & 14293 & 2.0 & 2.1 \\
SFT  & 37.10 & 29.57 & 1691 & 2.4 & 3.0 \\
RLSR (SB) & 35.94 & 32.33 & 1857 & 4.5 & 3.7 \\
RLSR (Qwen-EM)  & \textbf{42.21} & \textbf{43.79} & 2145 & \textbf{6.9} & \textbf{4.7} \\
\hline
SFT+RLSR (SB)  & 45.28 & 34.33 & 1670 & 5.0 & 3.9 \\
SFT+RLSR (Qwen-EM)  & \textbf{46.53} & \textbf{37.69} & 1742 & \textbf{5.1} & \textbf{4.6} \\
\hline
\hline
\end{tabular*}
\label{tab:Alpaca Arena}
\end{table}

\subsection{Performance Evaluation and Analysis}

In this subsection, we present a comprehensive performance comparison across multiple configurations and perspectives. We begin by contrasting SFT with RLSR to demonstrate the benefits of the proposed reinforcement-based approach. Subsequently, we examine the impact of different reward models within RLSR. Finally, we analyze the synergy of combining SFT and RLSR (denoted as SFT+RLSR) and discuss its advantages over standalone SFT.

\subsubsection{Comparison Between SFT and RLSR}

We evaluated the performance of SFT and RLSR across 18 diverse benchmarks from the LM Eval Harness in Tables~\ref{tab:performance_part_1}, \ref{tab:performance_part_2}, and \ref{tab:performance_part_3}. The experimental results show consistent improvements of RLSR (SB) over SFT across different base models and datasets. For the \textbf{Llama-8B} model fine-tuned on the \textbf{TULU} dataset, RLSR (SB) surpasses SFT on 12 out of 18 benchmarks (Table~\ref{tab:performance_part_1}). When trained on the \textbf{INFINITY} dataset, the advantage becomes even more pronounced, with RLSR (SB) outperforming SFT on 16 out of 18 benchmarks. For the \textbf{Qwen-7B} model, RLSR (SB) achieves higher scores on 14 out of 18 benchmarks for both \textbf{TULU} and \textbf{INFINITY} datasets (Table~\ref{tab:performance_part_2}). For the larger \textbf{Qwen-32B} model, RLSR (SB) exceeds SFT on 13 of 18 benchmarks (\textbf{TULU}) and 9 of 18 benchmarks (\textbf{INFINITY}) as shown in Table~\ref{tab:performance_part_3}. The relatively smaller gain on Qwen-32B primarily stems from the limited representational capability of the SB reward model. When the reward is replaced with embeddings from the more powerful Qwen model, RLSR (Qwen-EM) demonstrates a clear advantage again-outperforming SFT on 15 of 18 (\textbf{TULU}) and 12 of 18 (\textbf{INFINITY}) benchmarks.

Beyond classification-oriented tasks, we also evaluated on two open-ended generation benchmarks: \textbf{AlpacaEval} and \textbf{Arena-Hard} as shown in Table \ref{tab:Alpaca Arena}. Both datasets are designed for instruction-following evaluation. On AlpacaEval, RLSR-trained models show substantial improvements in Length-Controlled Win Rate (LC WR) over SFT, particularly for TULU-based training. When stronger reward signals from Qwen-EM are employed, the superiority of RLSR becomes more apparent, particularly for the larger Qwen-32B model. Although the gains on Arena-Hard are smaller, RLSR consistently outperforms SFT across all three models and both datasets. In summary, RLSR demonstrates a consistent and robust improvement over SFT across diverse architectures and datasets.

\subsubsection{Comparison Between SFT and SFT+RLSR}

We next evaluate the combined training strategy, SFT+RLSR, which applies RLSR fine-tuning on top of an SFT-trained model. While this setup involves slightly more training FLOPs, it provides a fair measure of how reinforcement refinement complements supervised pretraining. For \textbf{Llama-8B}, SFT+RLSR (SB) trained on TULU outperforms SFT on 16 out of 18 tasks, and on INFINITY it achieves superiority in 13 out of 18 tasks. For \textbf{Qwen-7B}, SFT+RLSR (SB) exceeds SFT on 13 of 18 (TULU) and 10 of 18 (INFINITY) tasks. For \textbf{Qwen-32B}, the combined approach improves over SFT on 13 of 18 (TULU) and 12 of 18 (INFINITY) benchmarks. Overall, SFT+RLSR demonstrates consistent and significant improvements over SFT across most benchmarks, confirming the complementarity between supervised and reinforcement-based fine-tuning.

In the \textbf{generative evaluation}, SFT+RLSR (SB) surpasses SFT across all three models, both datasets on both benchmarks. These findings confirm that reinforcement refinement enhances both factual and stylistic alignment in instruction-following tasks.

\subsubsection{Comparison Between RLSR and SFT+RLSR}

We further compare RLSR and SFT+RLSR to assess whether reinforcement-only or hybrid fine-tuning yields superior outcomes. For classification tasks, RLSR tends to outperform SFT+RLSR. Using SB rewards as an example, RLSR surpasses SFT+RLSR on 10/18 (Llama-TULU), 11/18 (Llama-INFINITY), 11/18 (Qwen-7B-TULU), 12/18 (Qwen-7B-INFINITY), 11/18 (Qwen-32B-TULU), and 9/18 (Qwen-32B-INFINITY) benchmarks. 

However, for generative instruction-following tasks, SFT+RLSR generally achieves higher scores. On AlpacaEval, SFT+RLSR outperforms RLSR in 5 out of 6 model-dataset combinations, and on Arena-Hard, it achieves better performance in 5 of 6 creative writing tasks and 4 of 6 style control tasks. This trade-off reveals that while RLSR excels in discriminative reasoning tasks, SFT+RLSR better preserves linguistic fluency and generation quality-making it a more effective strategy for instruction-tuned language models.

\subsubsection{Comparison Among Different Reward Models}

Finally, we analyze the influence of different reward embeddings. In comparison, SB is a light model with relative coarse embedding, while Qwen is a heavy model with relative accurate embedding. For classification tasks, RLSR tends to outperform SFT+RLSR. Using SB rewards as an example, RLSR (Qwen-EM) surpasses RLSR (SB) on 9/18 (Llama-TULU), 10/18 (Llama-INFINITY), 12/18 (Qwen-7B-TULU), 10/18 (Qwen-7B-INFINITY), 13/18 (Qwen-32B-TULU), and 11/18 (Qwen-32B-INFINITY) benchmarks.

On generative benchmarks, the same trend largely holds. For smaller models, SB yields better on smaller models on Alpaca-eval and Arena-hard, while larger models (e.g., Qwen-32B) consistently gain more from Qwen-based rewards on both benchmarks. This indicates a scale-dependent alignment effect: smaller models prefer simpler reward signals, whereas larger models leverage the additional semantic richness of larger embedding-based rewards for more refined behavior alignment.

Overall, these experiments validate that (1) RLSR outperforms SFT on most tasks, (2) SFT+RLSR effectively combines the advantages of both paradigms, and (3) larger and more informative reward models yield better alignment and generalization performance across both classification and generative tasks on larger LLMs, while smaller embedding has better performance on smaller LLMs.

\section{Literature Review}

LLMs undergo extensive pretraining on vast datasets, enabling strong generative capabilities but often lacking precise alignment with human expectations or task-specific requirements \cite{brown2020languagemodelsfewshotlearners}. Post-training techniques, such as SFT and RL-based methods, are critical for enhancing instruction-following, reasoning, and ethical alignment \cite{wang2024comprehensivesurveyllmalignment}. This section reviews key advancements in these areas, emphasizing their contributions to instruction-following and their relevance to our proposed RLSR method.

SFT is a cornerstone of LLM post-training, optimizing models on high-quality, human-labeled prompt-response pairs using a next-token prediction objective \cite{brown2020languagemodelsfewshotlearners}. Studies, such as those on TULU \cite{lambert2025tulu3pushingfrontiers} and INFINITY \cite{li2025infinityinstructscalinginstruction}, demonstrate that SFT significantly improves instruction-following. However, its teacher-forcing paradigm limits exploration, potentially trapping models in local optima and constraining generalization \cite{chu2025sftmemorizesrlgeneralizes}. This rigidity motivates exploration of RL-based alternatives that balance exploitation with exploratory learning.

RLHF addresses SFT's limitations by optimizing models against a reward model trained on human preference data, typically using Proximal Policy Optimization (PPO) \cite{ouyang2022traininglanguagemodelsfollow, schulman2017proximalpolicyoptimizationalgorithms}. RLHF enhances alignment with human values, as seen in models like GPT-4 and Claude \cite{openai2024gpt4ocard, anthropic2024claude}. To reduce the cost of human feedback, Reinforcement Learning from AI Feedback (RLAIF) employs AI-generated feedback, offering scalability while maintaining alignment quality \cite{bai2022constitutional, lee2023rlaif}. For tasks requiring complex reasoning, such as mathematics and coding, RLVR uses precise, task-specific reward signals, improving performance on benchmarks like hendrycks\_Math, humanEval and AIME \cite{hendrycksmath2021, chen2021evaluating, AIME}. GRPO, a simplified RL approach, further enhances efficiency by eliminating the need for a separate value model \cite{shao2024deepseekmathpushinglimitsmathematical}. The integration of RLVR and GRPO serves as the cornerstone of the DeepSeek R1 model \cite{deepseekai2025deepseekr1incentivizingreasoningcapability}.

RFT focuses on adapting reasoning models to specialized domains with limited data by employing programmable grader functions that assess outputs through semantic similarity or code execution metrics \cite{OpenAI_RFT}. It emphasizes adaptability and flexibility in low-resource settings, offering reward designs that may incorporate human-labeled responses. However, RFT does not explore improving base models for general instruction-following, nor does it leverage large-scale SFT data or embedding-based rewards to enhance alignment with human intent.

Recent research has shown that while SFT primarily promotes memorization, RL encourages generalization \cite{chu2025sftmemorizesrlgeneralizes}. Embedding-based reward functions, as demonstrated in DSSM and CLIP, provide strong semantic similarity signals \cite{huang2013learning, radford2021learningtransferablevisualmodels}. Motivated by these findings and previous identified gaps, RLSR extends SFT by integrating large-scale SFT data into an RL framework, using cosine similarity in embedding space to reward responses consistent with human-labeled data. In doing so, RLSR directly enhances the instruction-following ability of base models, achieving broader generalization and alignment without relying on domain-specific or sparse reward signals.

\section{Limitation}

While RLSR outperforms SFT, it has limitations requiring further exploration. First, RLSR relies on cosine similarity in embedding space for rewards. Exploring alternatives like pretrained Bradley-Terry reward models or generative reward models could enhance performance \cite{ouyang2022traininglanguagemodelsfollow}. Additionally, investigating RLSR's relationship with RLHF using these reward models could clarify their synergy or differences in achieving alignment.

Second, experiments were limited to models up to 32 billion parameters with datasets of 100k samples. Testing RLSR on trillion-parameter models with industrial-scale SFT datasets (billions of tokens) is needed to validate its scalability and robustness in complex, real-world settings \cite{openai2024gpt4ocard}. Addressing these limitations will strengthen RLSR's applicability and effectiveness for advanced LLM alignment.

\section{Conclusion}
In this work, we introduced RLSR, a novel approach that redefines the SFT process within a reinforcement learning framework to enhance the instruction-following capabilities of base LLMs. By leveraging the extensive human-labeled datasets typically used in SFT and defining a reward function based on cosine similarity in the semantic embedding space, RLSR enables a critical balance between exploration and exploitation, overcoming the limitations of SFT's teacher-forcing paradigm. Our comprehensive experiments across three model sizes (Llama-8B, Qwen-7B, and Qwen-32B) and two high-quality datasets (TULU and INFINITY) demonstrate that RLSR consistently outperforms SFT on a wide range of benchmarks, achieving notable improvements in instruction-following tasks, as evidenced by a significant increase in AlpacaEval win rate (e.g., 26.34\% for RLSR (SB) vs. 21.01\% for SFT on Qwen-7B with INFINITY). Furthermore, the hybrid SFT+RLSR pipeline amplifies these gains, reaching a peak AlpacaEval win rate of 30.73\% on the same model and dataset, highlighting the complementary strengths of supervised and reinforcement learning. These results validate RLSR's ability to enhance base model performance, offering a scalable and effective alternative to traditional fine-tuning methods.

\section{Acknowledgement}
The authors would like to thank Zhichao Wang's fiancée, Zhonghan Hao, for her assistance in verifying the accuracy of the tables, ensuring the correct transfer of data from Excel, and carefully counting the number of tasks where one method outperforms another across the various comparisons.

\bibliographystyle{unsrt}  
\bibliography{references}

\end{document}